\documentclass[article, letterpaper, 10 pt, conference]{ieeeconf}  

\IEEEoverridecommandlockouts                              

\usepackage{graphicx,subfig}

\usepackage{hyperref}
\usepackage[sorting=none, style=ieee]{biblatex}
\usepackage{booktabs}  
\usepackage{threeparttable} 
\usepackage{makecell}
\usepackage{graphicx}
\usepackage{multicol}
\usepackage{multirow}
\usepackage{feyn,soul}
\usepackage{amsmath,amssymb,amsfonts}
\usepackage{ulem}
\usepackage{caption}
\usepackage{fancyhdr} 
\usepackage{titlesec}
\usepackage{pifont}
\usepackage{amssymb}

\usepackage{hyperref}
\hypersetup{
	colorlinks=true,
	linkcolor=cyan,
	filecolor=blue,      
	urlcolor=black,
	citecolor=green,
}

\usepackage{biblatex}

\title{\LARGE \bf
Annotation-Free Curb Detection Leveraging Altitude Difference Image
}

\author{Fulong Ma, Peng Hou, Yuxuan Liu, Yang Liu, Ming Liu, and Jun Ma, \textit{Senior Member, IEEE}    
\thanks{Fulong Ma and Ming Liu are with The Hong Kong University of Science and Technology (Guangzhou), Guangzhou, China. (email: fmaaf@connect.hkust-gz.edu.cn, eelium@hkust-gz.edu.cn).}
\thanks{Peng Hou is with the Department of Electronic Engineering, Tsinghua University, Beijing, China. (email: houp18@mails.tsinghua.edu.cn)
}
\thanks{Yuxuan Liu, Yang Liu, and Jun Ma are with The Hong Kong University of Science and Technology, Hong Kong SAR, China. (email: yliuhb@connect.ust.hk, jun.ma@ust.hk).}%
}

\begin{document}

\maketitle

\thispagestyle{empty}
\pagestyle{empty}

\begin{abstract}


\footnotesize Road curbs are considered as one of the crucial and ubiquitous traffic features, which are essential for ensuring the safety of autonomous vehicles. 
Current methods for detecting curbs primarily rely on camera imagery or LiDAR point clouds.
Image-based methods are vulnerable to fluctuations in lighting conditions and exhibit poor robustness, while methods based on point clouds circumvent the issues associated with lighting variations. However, it is the typical case that significant processing delays are encountered due to the voluminous amount of 3D points contained in each frame of the point cloud data.
Furthermore, the inherently unstructured characteristics of point clouds poses challenges for integrating the latest deep learning advancements into point cloud data applications.
To address these issues, this work proposes an annotation-free curb detection method leveraging Altitude Difference Image (ADI) (as shown in Fig. \ref{adv_vs_rgb}), which effectively mitigates the aforementioned challenges. 
Given that methods based on deep learning generally demand extensive, manually annotated datasets, which are both expensive and labor-intensive to create, we present an Automatic Curb Annotator (ACA) module. This module utilizes a deterministic curb detection algorithm to automatically generate a vast quantity of training data. Consequently, it facilitates the training of the curb detection model without necessitating any manual annotation of data. 
Finally, by incorporating a post-processing module, we manage to achieve state-of-the-art results on the KITTI 3D curb dataset \cite{zhao2024curbnet} with considerably reduced processing delays compared to existing methods, which underscores the effectiveness of our approach in curb detection tasks. 
Our code and data will be open-sourced at: 
\href{https://sites.google.com/view/adi-curb-detection}{https://sites.google.com/view/adi-curb-detection}.

\end{abstract}

\section{INTRODUCTION}
 Road curbs are key elements in traffic scenes that partition the ground into drivable and non-drivable zones.
In this sense, road curb detection \cite{romero2021road} becomes one of the critical tasks in the field of machine vision, with wide application value in areas such as autonomous driving, advanced driver assistance system (ADAS), and other intelligent transportation systems. Essentially, accurate detection and identification of road curbs can be rather useful for vehicles in tasks like path planning, path following, and precise vehicle positioning. \cite{hata2014robust}, thereby enhancing driving safety and efficiency.

\begin{figure}[t]
    \setlength{\abovecaptionskip}{0pt}
    \setlength{\belowcaptionskip}{0pt}
    \centering
    \includegraphics[width=1.0\linewidth]{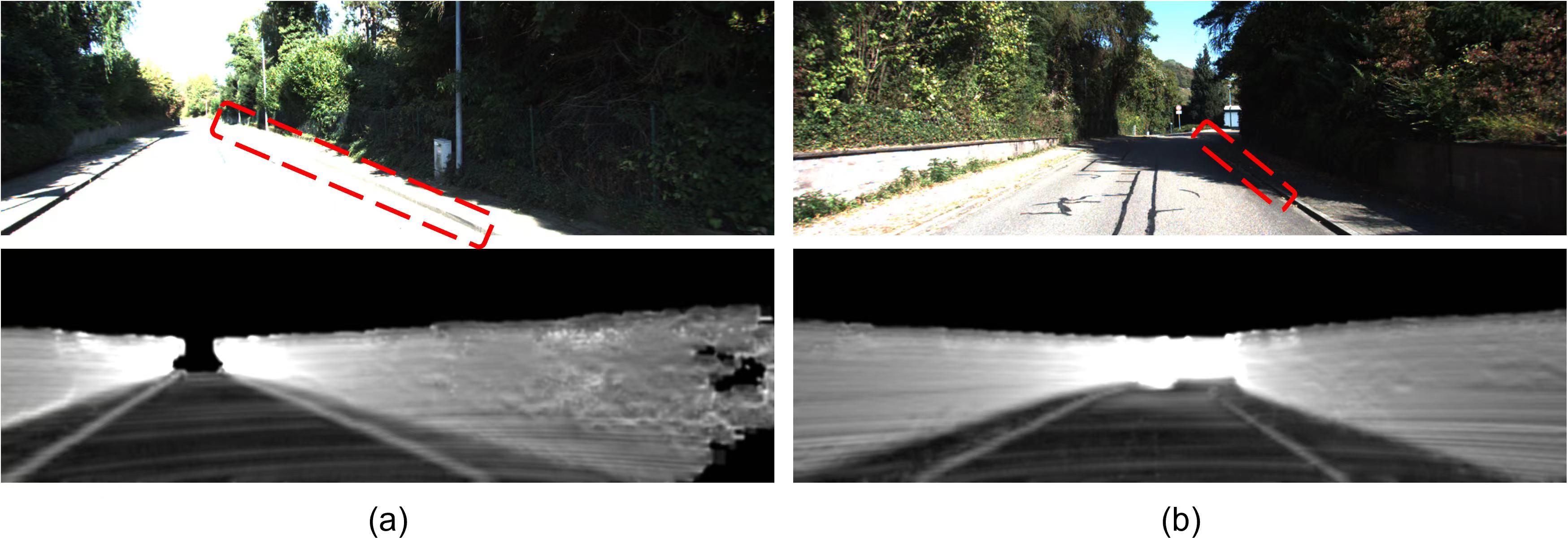}
    \captionsetup{font={footnotesize}}
    \caption{Comparison between RGB image and ADI under extreme lighting conditions. The top row represents RGB images, while the bottom row represents ADIs. In the first row, the image in column (a) is in a high-exposure scenario, while the image in column (b) is in a scenario with low lighting under tree shade. 
    It can be observed that under these extreme lighting conditions, the curb in the RGB images is difficult to discern (as indicated by the red dashed boxes). However, the curb is clearly visible in the ADIs.}
    \label{adv_vs_rgb}
\end{figure}

With the advancements in sensors and developments in deep learning technology, curb detection methods have become increasingly diverse and have made continuous progress.
From the perspective of sensor usage, curb detection can be classified into camera-based methods, LiDAR-based methods, ultrasonic-based methods, and multi-sensor fusion-based methods.
In camera-based methods, Romuald et al. used a monocular camera and a laser line scanner to detect curbs \cite{aufrere2003multiple}. In \cite{oniga2007curb,oniga2010polynomial}, researchers used stereo vision to detect road boundaries.
Currently, most road curb detection works are based on LiDAR sensors. LiDAR-based methods can be further subdivided into 2D LiDAR-based methods and 3D LiDAR-based methods. In terms of 2D LiDAR-based methods, curb detection can be achieved by finding the extreme points of lines crossing the road \cite{wijesoma2004road,liu2013curb}. In the context of 3D LiDAR-based methods, due to the high density of 3D LiDAR point clouds, advanced manual features such as curb height difference, smoothness, and horizontal distance are extracted to detect the curb \cite{wang2020speed,sun20193d,zhang2018road}. Alternatively, curb detection can be achieved through deep learning-based approaches \cite{zhao2024curbnet, zhou2020cylinder3d,gao2023lcdet}, which leverage an end-to-end architecture without the need for manually designed features.
Ultrasonic sensors operate by detecting the echo from the nearest obstacle, and their detection range typically varies from a few centimeters to several meters. Due to their extremely low cost, there have been a few studies attempting to use ultrasonic sensors for curb detection \cite{rhee2019low}, with the main drawbacks being lower accuracy and a more limited detection range.
Multi-sensor data fusion has been applied in various fields of autonomous driving, and there are also multi-modal fusion methods related to curb detection. In \cite{tan2014robust}, it fused LiDAR and images to create a depth image, and then the depth image was used for curb detection. In  \cite{goga2018fusing,deac2019curb}, image data and data from multiple LiDAR sensors were first fused, and then the data was fed into classification algorithms to address the curb detection problem.

While deep learning advancements have greatly enhanced curb detection, the challenge of obtaining high-quality training data remains a persistent issue in the field.
The annotation of training data is an extremely time-consuming, labor-intensive, and costly task. Due to the high cost of manual data annotation, it has somewhat hindered the application of deep learning in the field of autonomous driving.
To promote the application of deep learning in the field of autonomous driving, researchers have been continuously exploring low-cost methods for obtaining training data. For example, in  \cite{wang2019self,ma2023self}, the authors utilized depth information from stereo vision and LiDAR to automate the generation of training data for free space detection tasks. In \cite{ai2023ws,nie2024lanecorrect}, researchers proposed a weakly-supervised and self-supervised 3D lane detection method by leveraging the assumption of equal lane widths and heights for adjacent lanes, as well as the reflective characteristics of the materials used in lane marking paints.
However, due to the lack of this reflective property in curbs, this method is not suitable for curb detection tasks.

In summary, the curb detection task is confronted with various challenges. Firstly, RGB images are susceptible to lighting conditions, while point cloud data is excessive but leading to significant delays in processing. Also, unstructured data is not ideal for deep learning models. Secondly, the costly manual data annotation process hampers the practical implementation of curb detection technologies.
To address these challenges, we propose an annotation-free curb detection method based on Altitude Difference Images (ADIs)  By converting point clouds into ADIs, We address the data-related challenges associated with the time-consuming processing of point clouds, and also the sensitivity of RGB images to changes in lighting conditions.
Furthermore, with the implementation of an Automatic Curb Annotator (ACA) module, we notably diminish the reliance on manually annotated data. To the best knowledge of the authors, our
development is the first annotation-free approach for curb
detection tasks. 

In summary, our contributions are as follows:
\begin{itemize}


 \item We introduced a novel curb detection technique with ADIs converted from point clouds, which is shown to be more resilient to variations in lighting conditions compared to RGB-based methods. Also, it offers a more streamlined and computationally efficient pipeline compared to point cloud-based approaches.
 
\item We proposed an automatic curb annotator module that  automatically generates training data, and this eliminates the need for manual data annotation and significantly reduces the cost of model training.

\item We conducted extensive experiments on the KITTI 3D curb dataset, and the results indicated that our proposed method achieved state-of-the-art performance while notably reducing processing latency. This further highlights its suitability for deployment in real-world applications that require the real-time performance.


\end{itemize}

\section{RELATED WORKS}

\subsection{Curb Detection}

LiDAR and camera are the most commonly used sensors for curb detection, and there is also very little work based on ultrasonic sensors \cite{rhee2019low}. Based on the data from these sensors, curb detection methods are generally divided into traditional methods and learning-based methods.

In the early stages, Romuald \textit{et al.} used a monocular camera with the Canny edge detection algorithm \cite{ding2001canny} to obtain the region of interest and then combined it with a laser line scanner to detect the position information of the curb \cite{aufrere2003multiple}.
Subsequently, a series of works were developed based on stereo vision~\cite{oniga2010polynomial,oniga2007curb}, which derived depth information from binocular sensors, and then extracted the curb based on this depth information. However, such methods have limited effectiveness due to the precision constraints of depth information from binocular sensors and the inherent time-consuming nature of stereo matching algorithms.
Subsequently, the emergence of LiDAR brought new solutions and progress to this field.
In the context of LiDAR-based methods, Zhang \textit{et al.} \cite{zhang2018road} proposed a curb detection approach based on road segmentation. In \cite{sun20193d}, a GNSS system was integrated to search for candidate curb points based on the vehicle's trajectory, and then geometric features of the curb points were used to filter out the candidates to obtain the final curbs. In \cite{wang2020speed}, Wang \textit{et al.} extracted three features: height difference, smoothness, and horizontal distance to detect curbs, such that a good trade-off between real-time performance and effectiveness was achieved. While these methods have made significant progress in the existing literature, their drawbacks are also readily evident, such as the requirement for manual feature design and frequent parameter tuning in response to scene changes.

On the other hand, learning-based curb detection methods have also attracted considerable attentions. 
Cheng \textit{et al.} \cite{cheng2018curb} designed a 16-dimensional descriptor based on the appearance, geometry, and disparity features of curbs, using disparity and v-disparity maps generated from stereo matching, and employed an SVM classifier to detect curbs. However, this method involves too many intermediate steps, resulting in low efficiency.
In \cite{jung2021uncertainty}, researchers proposed an uncertainty-aware convolutional neural network that detects curbs by predicting them through autoencoders, and then refines the predictions with uncertainty estimations using conditional neural processes. However, when converting point clouds to the bird's-eye view (BEV), the sparse curb features may be lost, especially with low-beam LiDAR.
Gao \textit{et al.} proposed LCDeT \cite{gao2023lcdet}, a transformer-based architecture for end-to-end extraction of curbs from laser scan point clouds. However, there is a need to improve the computational efficiency of networks that leverage the transformer architecture.
Moreover, if curb detection is treated as a semantic segmentation task, then general point cloud semantic segmentation models such as \cite{wu2018squeezeseg,zhou2020cylinder3d,hou2022point,milioto2019rangenet++} can be utilized for curb detection tasks. It should be pointed out that, since general semantic segmentation models are designed for full-scene segmentation, class imbalance is a common issue in curb detection tasks, given the smaller size of curbs compared to the background. This necessitates targeted improvements to the general segmentation models to address the issue.

\subsection{Annotation-Free Approaches in Deep Learning}
It is widely recognized that in deep learning, the quality of the data is just as important as the deep learning models themselves.
Nevertheless, the high cost associated with acquiring manually annotated data poses a challenge and can limit the advancement of deep learning techniques.
In \cite{he2020momentum,caron2021emerging,he2022masked}, researchers have proposed self-supervised training methods without the need for data annotation to train the backbone network, thereby enhancing the performance of downstream tasks. However, these methods can only transfer certain shared features to downstream tasks, and annotated data on downstream tasks is still required to fine-tune the model.
In the context of task-specific self-supervised methods, 
Ma \textit{et al.} \cite{ma2019self} proposed a self-supervised method using images to achieve sparse-to-dense depth completion. 
In \cite{ma2019self}, an annotation-free drivable area detection method was proposed based on images with the assistance of point clouds. Also, Nie \textit{et al.} proposed LaneCorrect  \cite{nie2024lanecorrect}, which deals with the self-supervised lane detection problem by automatically correcting lane labels through learning geometric consistency and instance awareness from adversarial augmentations. 
In recent years, with the development of foundation models, a series of works, such as \cite{li2024clipsam,zhou2023zegclip},  leveraged the strong generalization capabilities of these foundation models to achieve excellent performance on downstream tasks without the need for any data annotation.
Nevertheless, while promising in the context of annotation-free approaches for deep learning, a more concrete strategy and pipeline is still lacking on how to extend the annotation-free concept in curb detection tasks.

\section{METHOD}

\subsection{LiDAR Point Cloud to Altitude Difference Image}
The ADI was first introduced in PLARD \cite{chen2019progressive}. When calculating the ADI, the $z$ channel of the LiDAR is treated as the altitude, and the ADI is obtained by calculating the differences in the $z$ channel values. 
Specifically, the altitude difference-based transformation computes the value at the pixel coordinates $(x, y)$ as:
\begin{equation}
    f(x,y) = \frac{1}{M} \sum_{N_{x},N_{y}} \frac{| Z_{x,y} - Z_{N_{x},N_{y}}|}{\sqrt{(N_{x}-x)^{2} + (N_{y} - y)^{2}}},
\end{equation}
where $Z_{x,y}$ is the altitude of the LiDAR point projected on $(x, y)$, $(N_{x},N_{y})$ represents positions in the neighborhood, $M$ is the total number of the considered neighborhood positions.
Through this transformation, the LiDAR point cloud is converted into an image format with a regular data structure. Objects with altitude differences, such as curbs, exhibit prominent features in the ADI. Since the ADI is derived from LiDAR point clouds, which are unaffected by lighting conditions, the resulting ADI is highly robust to changes in illumination, as demonstrated in Fig. \ref{adv_vs_rgb}.
\subsection{Automatic Curb Annotator}

\subsubsection{LiDAR-Based Road Curb Detection}

We employ the road curb detection algorithm proposed in Wang \textit{et al.} \cite{wang2020speed} to extract road information. The method consists of four steps: ground segmentation, feature extraction, feature classification, and feature filtering. A brief introduction to each step is provided below:

\textit{(a) Ground Segmentation:} To improve the modeling of the ground surface, we employ a piecewise plane fitting method to classify ground points and non-ground points. Represented by $P$, a frame of raw point cloud is first divided into multiple segments along the $x$-axis, and then a plane fitting algorithm \cite{zermas2017fast} is applied in each segment to extract ground points $P_g$ and non-ground points $P_{ng}$.

\textit{(b) Feature Extraction:}
To extract feature points, three spatial features are employed. The ground points $P_g$ are divided into the same number of scan layers as the sensor rings for feature extraction, with each layer comprising all points from the corresponding laser. Let $P_{ri} = [x_{ri},y_{ri},z_{ri}]$ represent each point, with $r$ denoting the corresponding ring ID. Then, the three spatial features are outlined as follows:

\begin{itemize}
    \item {Height Difference:}
    Let $Z_{max}$ and $Z_{min}$ denote the maximum and minimum of $z$ values of neighbors of $P_{ri}$, respectively. Then, the height difference feature is defined as:
    \begin{equation}
        H_{1} \leq Z_{max} - Z_{min} \leq H_{2},
    \end{equation}
    \begin{equation}
        \sqrt{\frac{\sum(z_{ri} - \mu)^{2}}{n_{neighbor}}} \geq H_{3},
    \end{equation}
    where $n_{neighbor}$ is the number of neighbors, $z_{ri}$ is the $z$ value of each neighbor, $H_{1}$, $H_{2}$, and $H_{3}$ are thresholds. Also, $\mu =\sum z_{ri} / n_{neighbor}$.
   
    \item {Smoothness:}
     This feature is defined as:
     \begin{equation}
         s = \frac{\big \| \sum_{P_{ri} \in S, j \neq i} (p_{ri} - p_{rj})  \big \|}{|S|\cdot||P_{ri}||},
     \end{equation}
     \begin{equation}
         s \geq T_{s},
     \end{equation}
     where $s$ is the smoothness value of $p_{ri}$, $|S|$ is the cardinality of $S$, and $T_{s}$ is the threshold.
    \item {Horizontal Distance:}
    This feature is defined as:
    \begin{equation}
        \delta_{xy, r} = H_{s} \cdot \cot \theta_{r} \cdot \pi\theta_{a},
    \end{equation}
    where $H_{s}$ is the absolute value of the height of point $p_{ri}$, $\theta_{l}$ is the vertical azimuth of scanning layer $r$, $\theta_{a}$ is the horizontal angular resolution of LiDAR. Note that $\theta_{r}$ and $\theta_{a}$ are both in radians.
\end{itemize}

\textit{(c) Feature Classification:}
A road-segmentation-line-based method is used for classifying feature points, where the principle is based on establishing a beam model \cite{thrun2002probabilistic} from non-ground points $P_{ng}$, from which the length of each beam is calculated. The longest beam in the front and rear regions of the vehicle are considered as road segmentation line to indicate the direction of the road. Additionally, a peak-finding algorithm is employed to filter out outliers and determine two truly dominant extremes in the distance function. Finally, feature points are divided into left and right feature points based on the road segmentation line.

\textit{(d) Feature Filtering:}
After extracting and classifying feature points, there are still some false points present. In order to filter out these false points, an iterative Gaussian Process Regression algorithm \cite{schulz2018tutorial} is used in the feature point filtering step to model road boundaries and filter out false points.

\begin{figure*}[t]
    \setlength{\abovecaptionskip}{0pt}
    \setlength{\belowcaptionskip}{0pt}
    \centering
    \includegraphics[width=1.0\linewidth]{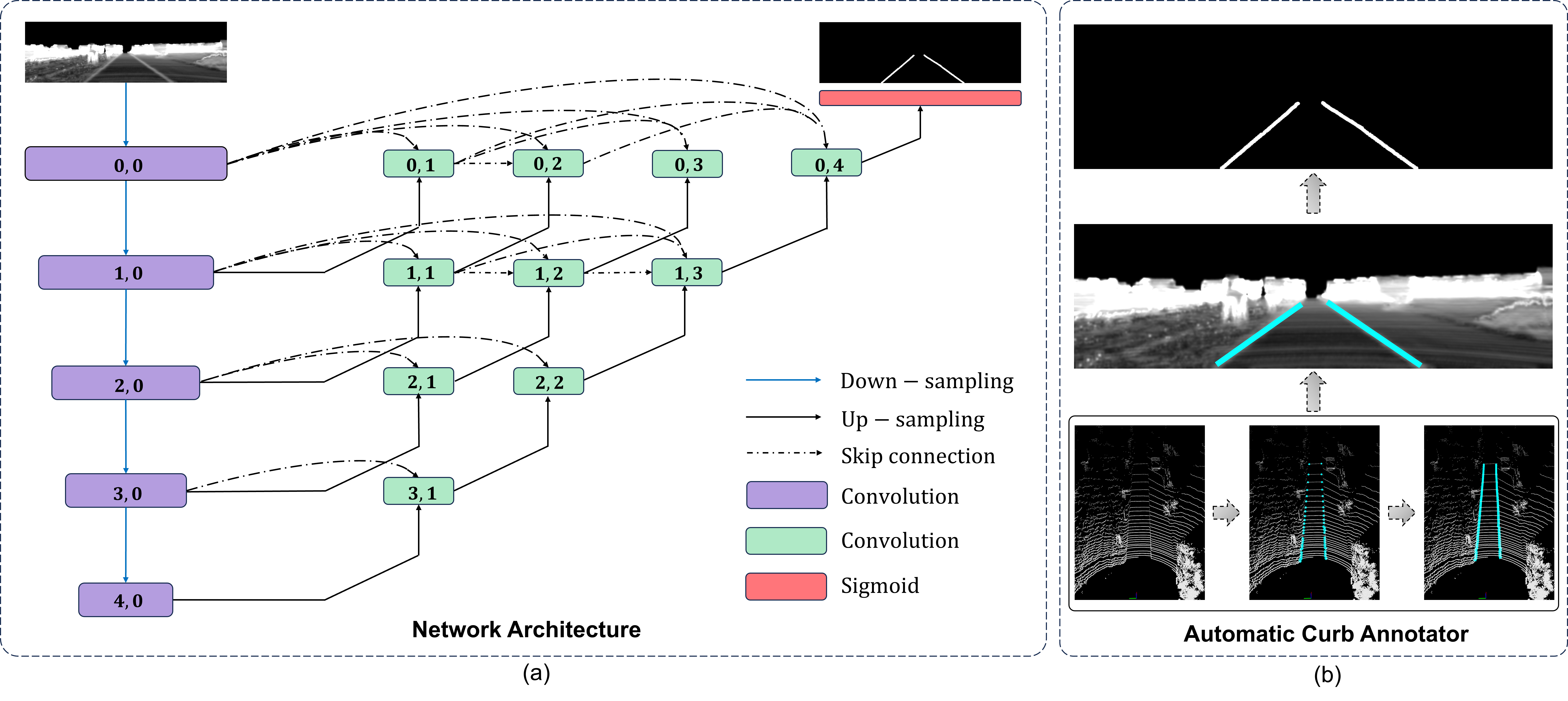}
    \captionsetup{font={footnotesize}}
    \caption{The overall framework of our proposed curb detection method based on ADIs with an annotation-free approach. (a): Schematic diagram of the network architecture. (b): Illustration of the process in the ACA module.
}
    \label{pipeline}
\end{figure*}

In addition to these four steps, we further add a step to remove non-fixed-scene object point clouds in order to reduce the adverse effects of vehicles and other objects on road curb detection. 

\subsubsection{Projection of 3D Curb Points to 2D Image Plane}
We use the pinhole model to project the 3D points onto the 2D image plane. 
The projection of a 3D point $ \textbf{x} = (x, y, z, 1)^{T} $ in rectified (rotated) camera coordinates to a point $ \textbf{y} = (u, v, 1)^{T} $ in the camera image plane is given as $\textbf{y} = \textbf{P} \cdot \textbf{x}$,
with

\begin{equation}  \label{eq2}
    \textbf{P} =
	\begin{pmatrix} f_{u}  & 0     & c_{u} & -f_{u}b_{x} 
                      \\ 0 & f_{v} & c_{v} & 0 
                      \\ 0 & 0     & 1     & 0     
    \end{pmatrix}.
\end{equation}
Here, $f_{u}$, $f_{v}$ are focal length in pixels, $c_{u}$, $c_{v}$ are optical center in pixels, $b_{x}$ denotes the baseline with respect to reference camera.
In this way, we project the road curb points in the LiDAR coordinate to the image plane. 
The LiDAR-based road curb detection module and the 3D curb points to 2D image plane projection module together constitute our ACA module, and a schematic diagram of the ACA module is shown in Fig. \ref{pipeline}.

\subsection{Extremely Lightweight Backbone}
Autonomous driving and ADAS systems have extremely high requirements for real-time performance, as excessive computational latency may lead to catastrophic consequences.
Moreover, for autonomous driving and ADAS systems, the onboard processors also run other algorithms, such as localization, object detection, and planning. Therefore, for the individual task of curb detection, it is essential to minimize its processing latency as much as possible to ensure the real-time performance of the entire system.
Therefore, in this paper, we choose MobileOne \cite{vasu2023mobileone} as our backbone network.

\begin{figure}[t]
    \setlength{\abovecaptionskip}{0pt}
    \setlength{\belowcaptionskip}{0pt}
    \centering
    \includegraphics[width=1.0\linewidth]{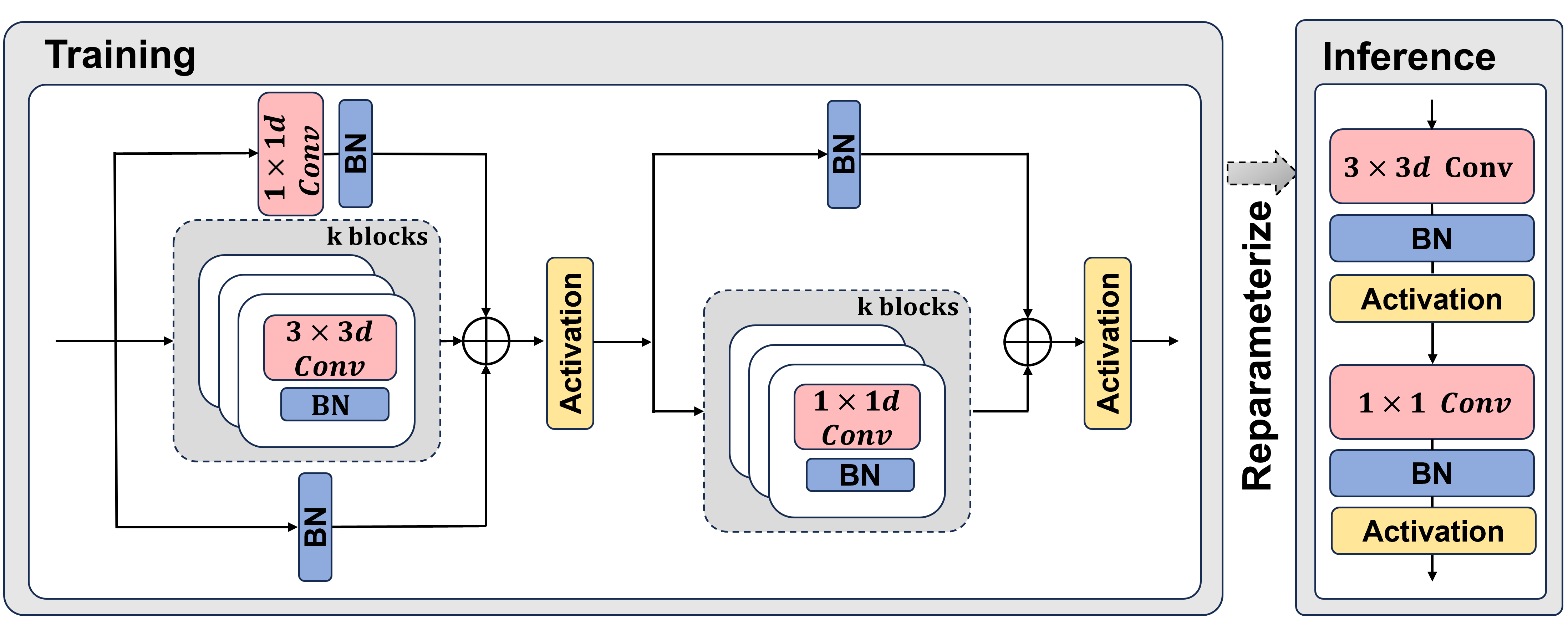}
    \captionsetup{font={footnotesize}}
    \caption{The MobileOne module has two different structures during training and inference. Left: MobileOne module during training with reparameterizable branches. Right: MobileOne module during inference where the branches are reparameterized.
}
    \label{mobileone_block}
\end{figure}

The basic block of MobileOne is designed based on MobileNet-V1 \cite{howard2017mobilenets}, utilizing the core architecture of a 3$\times$3 depth-wise convolution plus a 1$\times$1 point-wise convolution. During training, MobileOne applies the structural re-parameterization technique found in RepVGG \cite{ding2021repvgg}, by adding several parallel branches to both the 3$\times$3 depth-wise convolution and the 1x1 point-wise convolution, as illustrated in Fig. \ref{mobileone_block}. However, MobileOne's approach essentially differs from RepVGG: During training, RepVGG adds a 1$\times$1 convolution and a shortcut branch that only contains batch normalization (BN) to the 3$\times$3 convolution. In contrast, MobileOne adds $k$ 3$\times$3 depth-wise convolutions, a 1$\times$1 depth-wise convolution, and a BN-only shortcut branch to the 3$\times$3 depth-wise convolution. At inference time, the MobileOne model does not have any branches, which results in a streamlined architecture and further accelerates the model. The detailed practice of structural re-parameterization is aligned with that of RepVGG.

For the segmentation head, we used the UNet++ \cite{zhou2018unet++} architecture, which is an enhancement of the original UNet \cite{ronneberger2015u} architecture primarily focusing on optimizing the skip connections. In the UNet, directly combining shallow features from the encoder with deep features from the decoder might result in a semantic gap. To address this issue, researchers introduced a nested, dense skip connection scheme to optimize the feature fusion process, reducing the semantic gap between the feature maps of the encoder and decoder. This strategy facilitates the optimization process for the optimizer by enabling a more extensive and multi-scale feature integration. To this end, the overall curb detection network is depicted in the Network Architecture part of Fig. \ref{pipeline}.

\subsection{Post-Processing}

In practical applications, it is difficult to directly use the lane detection results based on perspective view (PV) for decision-making of autonomous vehicles.  
Therefore, we adopt the method from the KITTI road dataset \cite{fritsch2013new}, to transform PV images to a bird's-eye view (BEV) for evaluation.
However, when a PV image is converted to a BEV image, the image undergoes distortion because curbs in real-world scenarios do not perfectly lie on a horizontal plane, making the assumption of a flat surface in Inverse Perspective Mapping (IPM) to be invalid. Therefore, we need post-processing to refine the lane detection results in BEV.



In order to accelerate the processing time of post-processing as much as possible, we propose a two-stage processing method. The first step is to select candidate points, and the second step is to fit curves based on the candidate points.
Due to the deformation of curbs caused by IPM transformation, real curbs are generally located closer to the middle of the deformation area. Therefore, when selecting candidate points, we take the middle value of each curb pixel in the BEV to obtain curb candidate points. At this point, the curb candidate points contain noises and are not smooth, as shown in Fig. 4(d). Next, we fit the candidate points with quadratic curves to obtain a final result of curbs that is more accurate and smoother, as shown in Fig. 4(e). 
By using this method of first selecting candidate points and then fitting them, the number of points to be fitted has been reduced from approximately 30,000 to no more than 1,600 (BEV image height 800 pixels), a reduction of over 90\%, significantly reducing post-processing time.

\begin{figure}[t]
    \setlength{\abovecaptionskip}{0pt}
    \setlength{\belowcaptionskip}{0pt}
    \centering
    \includegraphics[width=1.0\linewidth]{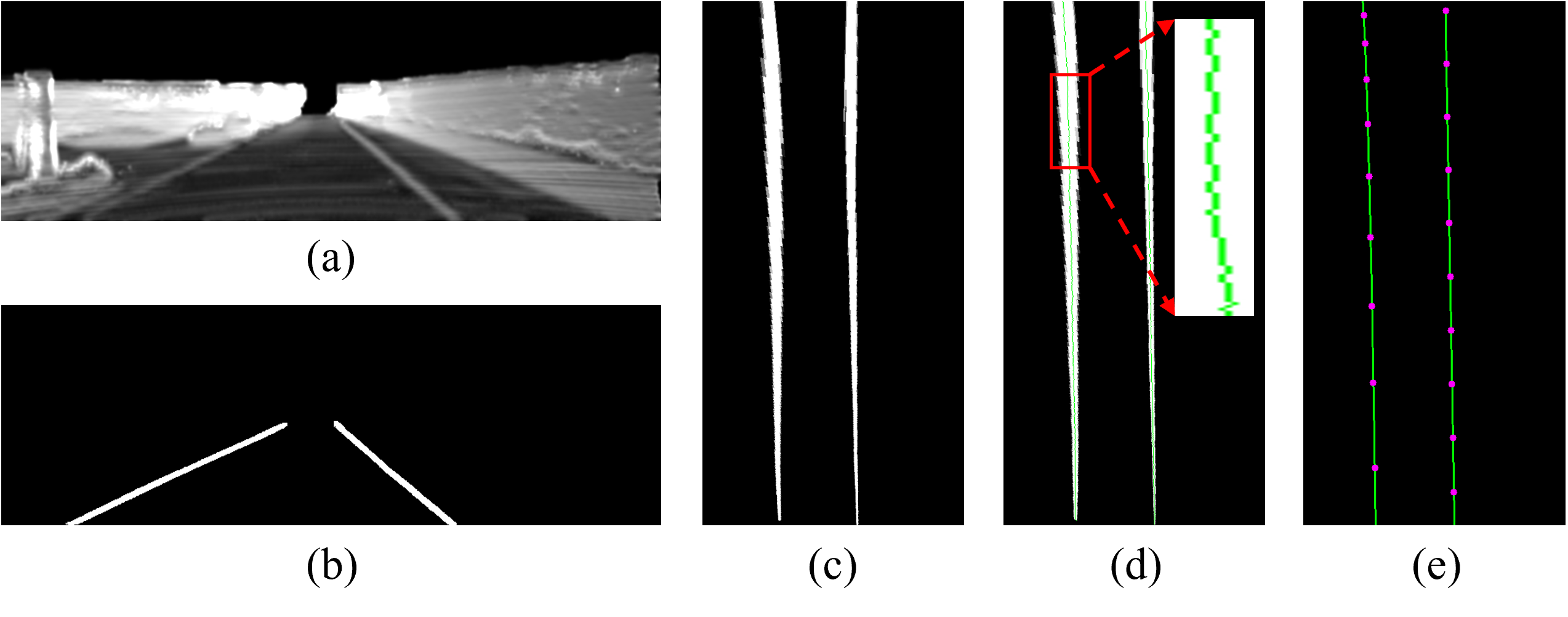}
    \captionsetup{font={footnotesize}}
    \caption{Data flow throughout the entire process from inputting the ADI into the network to obtaining the final curb detection results.
}
    \label{post_process}
\end{figure}

The data flow throughout the entire process
from inputting the ADI into the network to obtaining the final curb
detection results is shown in Fig. \ref{post_process}. Specifically, Fig. 4(a) is the input ADI, Fig. 4(b) is the network output result, Fig. 4(c) represents the BEV image obtained by transforming Fig. 4(b) through IPM, Fig. 4(d) represents the candidate points selected from Fig. 4(c), and Fig. 4(e) shows final curb detection result after all post-processing steps.

\begin{table*}[h]
\renewcommand\arraystretch{1.0}
\captionsetup{font={footnotesize}}
\caption{Comparison of our method with other methods on the KITTI 3D curb dataset \cite{zhao2024curbnet}.
}
\centering
\setlength{\tabcolsep}{4.0mm}

\begin{tabular}{c|ccccccc}
\hline Category & Method & Input & Annotation Free & Precision & Recall & F1-Score & FPS\\
\hline \multirow{3}{*}{\makecell{Traditional \\ Method}} & Zhang \textit{et al.} \cite{zhang2018road} &Point Cloud & N/A
                           & 84.90 & 82.88 & 83.88 & 20\\
& Sun \textit{et al.} \cite{sun20193d} &Point Cloud & N/A  & 87.36 & 84.27 & 85.79 & 33 \\
& Wang \textit{et al.} \cite{wang2020speed}&Point Cloud & N/A & 90.77 & 85.46 & 88.04 & 16 \\
\hline
\multirow{7}{*}{\makecell{Learning-based \\ Method}} 

& RangeNet++ \cite{milioto2019rangenet++} & Point Cloud&$\times$
                           &89.63  & 88.67 & 89.15 & 35 \\
& Cylinder3D \cite{zhou2020cylinder3d} & Point Cloud&$\times$
                           & 91.17 & 90.78 & 90.47 &16 \\
& PVKD \cite{hou2022point}&Point Cloud &$\times$ & 92.98 & 91.21 & 92.09 & 27 \\
& CurbNet \cite{zhao2024curbnet} & Point Cloud & $\times$
                           & 95.67 & 93.95 & 94.80 & 23\\
\cline{2-8}
& ADICurb-S0      &   ADI     & $\checkmark$  &94.14     &93.77      &93.83     &  \textbf{101}   \\
& ADICurb-S2       &   ADI    & $\checkmark$  & 95.88    &  95.87    &  95.84   &   83  \\
& ADICurb-S4      &   ADI   & $\checkmark$  &  \textbf{96.43}  & \textbf{96.59}    & \textbf{96.51}    & 58    \\
\hline
\end{tabular}
\label{results}
\end{table*}

\begin{table}[t!]
\renewcommand\arraystretch{1.0}
\captionsetup{font={footnotesize}}
\caption{The experimental results when the input is RGB image.
}
\centering
\setlength{\tabcolsep}{2.0mm}
\footnotesize
\begin{tabular}{ccccc}
\hline 
 Model &Input & Precision & Recall & F1-Score \\
\hline
ADICurb-S0 & RGB  & 82.34 & 84.76  & 83.53 \\
\hline
ADICurb-S2 & RGB  & 85.32 & 88.05  & 86.66  \\
\hline
ADICurb-S4 & RGB  &87.26  & 88.50  &  87.88 \\
\hline
\end{tabular}
\label{rgb_input}
\vspace{-15pt}
\end{table}

\section{EXPERIMENT}

\subsection{Experimental Setup}

\subsubsection{Baselines}
Although we perform curb detection on ADI, ADI is derived from LiDAR point cloud data. Essentially, it is pertinent to note that our method is still based on point clouds. Therefore, in this paper, we only compare it with point cloud-based methods.
We compared our method with both traditional methods and learning-based methods. 
Among the traditional methods, we compared several of the latest top-performing methods \cite{wang2020speed,zhang2018road,sun20193d}.
In terms of deep learning-based methods, we initially selected approach dedicated to curb detection, such as CurbNet \cite{zhao2024curbnet}, followed by the implementation of several baselines using representative architectures for semantic segmentation \cite{milioto2019rangenet++,zhou2020cylinder3d,hou2022point}. These architectures can be effectively utilized for LiDAR-based detection tasks, including the curb detection task.

\subsubsection{Dataset}
In this work, we used the KITTI 3D curb dataset 
\cite{zhao2024curbnet}, which is based on the Semantic KITTI dataset \cite{behley2019semantickitti}. It comprises a total of 7100 frames, with the training, validation, and testing sets consisting of 4300 frames, 1400 frames, and 1400 frames, respectively.

\subsubsection{Experiment Details}
All our experiments are conducted on a Ubuntu 20.04 system, equipped with a 12th Gen Intel i7-12700F CPU and a NVIDIA GeForce RTX 4090 GPU.
We employ PyTorch framework to train the models, setting the batch size to 32 for ADICurb-S0 and ADICurb-S2, and to 16 for ADICurb-S4. 
All models are trained for 300 epochs using the Adam optimizer with an initial learning rate of \(2 \times 10^{-4}\).

\subsection{Evaluation Metrics}

In alignment with other research works on road curb detection \cite{wang2020speed,zhao2024curbnet}, we used three common metrics to evaluate the performance of our proposed method. These evaluation metrics are: $Precision$, $Recall$, and $F1-Score$,
and they were computed as: $Precision =  \frac{N_{TP}}{N_{TP}+N_{FP}} , Recall =  \frac{N_{TP}}{N_{TP} + N_{FN}}, F1-Score =  \frac{2 \times Precision \times Recall}{Precision + Recall},$
where $N_{TP}$, $N_{TN} $, $N_{FP}$, and $N_{FN}$ represent the true positive, true negative, false positive, and false negative pixel numbers, respectively. We used a tolerance of 2 pixels, which means that if the detected curb is within 10 centimeters of the ground truth value, it is considered a true positive.

\subsection{Performance Evaluation}

We conducted experiments using the current best traditional methods and also the state-of-the-art learning-based methods for curb detection.  The comparative experimental results are shown in Table \ref{results}. 
To ensure a fair comparison, we standardize the evaluation approach for all methods used in this paper. That is, we convert the detection results of all methods and the ground truth to BEV for comparative analysis. Our method is essentially based on point cloud, but the point cloud is first converted into ADI, hence we denoted our input as ADI to differentiate it from other methods.


As shown in Table \ref{results}, we implemented three methods, namely ADICurb-S0, ADICurb-S2, and ADICurb-S4, which differ in that they respectively use MobileOne-S0, MobileOne-S2, and MobileOne-S4 as the backbone.
Our ADICurb-S0 has slightly lower metrics compared to CurbNet, but it achieves the fastest inference speed of 101 FPS, which is significantly higher than other methods.
Our ADICurb-S2 outperforms CurbNet in terms of metrics and operates at a speed of 83 FPS. Although the running speed is not as fast as ADICurb-S0, it is still significantly faster than other existing methods.
Our ADICurb-S4 achieves the best metrics. Specifically, compared to CurbNet, the precision is improved from 95.67 to 96.43, the recall is enhanced from 93.95 to 96.59, and the F1-Score is increased from 94.8 to 96.51. ADICurb-S4 operates at a speed of 58 FPS, which is significantly slower than ADICurb-S0 and ADICurb-S2, yet it remains faster than other existing methods. We also conduct experiments with RGB images as input to verify the superiority of using ADI, as shown in Table \ref{rgb_input}. With all other experimental settings being identical, the performance of ADICurb-S0, ADICurb-S2, and ADICurb-S4 all notably decrease when RGB images are used as input. We believe this is primarily due to the fact that in RGB images, the curb color features are often similar to the road surface in many cases, and RGB images are easily affected by exposure or insufficient lighting.

From these experimental results, it is evident that our method, despite not requiring manual data annotation, achieves excellent performance with fast operational speed. 
We attribute this primarily to the conversion of point cloud to ADI, which renders the curb features distinctly visible in the ADI, thereby facilitating the detection of curbs by deep neural network models. 
The structured data characteristics of ADI also make it easier to employ lightweight network architectures for acceleration purposes. 
For more detailed statistics on processing time, as shown in Table \ref{time_consumption}. Pre-processing refers to converting point cloud into ADI. Since only the point cloud within the range of the RGB image perspective are used, we cropped the original point cloud to retain only the point cloud within the range of the RGB image perspective. Both pre-processing and post-processing are implemented in C++, with each taking approximately 3 ms. The inference times of the deep neural network models under different backbone networks are 4 ms, 6 ms, and 11 ms, respectively. The overall processing times are 10 ms, 12 ms, and 17 ms, respectively.





\begin{table}[t!]
\renewcommand\arraystretch{1.0}
\captionsetup{font={footnotesize}}
\caption{Analysis of time consumption with our method.}
\centering
\setlength{\tabcolsep}{1mm}
\footnotesize
\begin{tabular}{ccccc}
\hline 
 Model &Pre-Processing & Model Inference & Post-Processing & Total \\
\hline
ADICurb-S0 &  3 ms & 4 ms & 3 ms  & 10 ms  \\
\hline
ADICurb-S2 &  3 ms & 6 ms & 3 ms  & 12 ms  \\
\hline
ADICurb-S4 &  3 ms & 11 ms & 3 ms  & 17 ms  \\
\hline
\end{tabular}
\label{time_consumption}
\vspace{-15pt}
\end{table}

\section{CONCLUSIONS}
In this work, we propose an annotation-free curb detection method based on ADIs. By converting LiDAR point clouds into ADIs, elements such as curbs, which exhibit a height difference from the ground, become more prominent, facilitating easier recognition by models. Additionally, since ADIs are derived from LiDAR point clouds, they exhibit high robustness against variations in lighting conditions. 
Another advantage of using ADIs is that they enable the application of the latest deep learning advancements, which are predominantly developed for structured data like images. By introducing the ACA module, our curb detection model can be trained without the need for manual annotations. Experimental results on the KITTI 3D curb dataset demonstrate that our method achieves state-of-the-art performance, while significantly reducing processing latency compared to existing methods, without requiring manual data annotation.
\label{conclusions}

\normalem
\renewcommand*{\bibfont}{\footnotesize}
\printbibliography




\end{document}